\documentclass[letterpaper]{article} % DO NOT CHANGE THIS
\usepackage{aaai24} % DO NOT CHANGE THIS
\usepackage{times} % DO NOT CHANGE THIS
\usepackage{helvet} % DO NOT CHANGE THIS
\usepackage{courier} % DO NOT CHANGE THIS
\usepackage[hyphens]{url} % DO NOT CHANGE THIS
\usepackage{graphicx} % DO NOT CHANGE THIS
\usepackage{natbib} % DO NOT CHANGE THIS AND DO NOT ADD ANY OPTIONS TO IT
\usepackage{caption} % DO NOT CHANGE THIS AND DO NOT ADD ANY OPTIONS TO IT
\usepackage{algorithm}
\usepackage{algorithmic}

\usepackage{newfloat}
\usepackage{listings}
\DeclareCaptionStyle{ruled}{labelfont=normalfont,labelsep=colon,strut=off} % DO NOT CHANGE THIS
\floatstyle{ruled}
\newfloat{listing}{tb}{lst}{}
\floatname{listing}{Listing}
\usepackage{amsmath}
\usepackage{amsthm}
\usepackage{amsfonts}

\usepackage{xcolor}
\usepackage{nicefrac}
\usepackage{cleveref}

\usepackage{subcaption}

\title{Who Reviews The Reviewers? A Multi-Level Jury Problem}
\author {
    Ben Abramowitz\textsuperscript{\rm 1},
    Omer Lev\textsuperscript{\rm 2},
    Nicholas Mattei\textsuperscript{\rm 1}
}
\affiliations {
    \textsuperscript{\rm 1}Tulane University\\
    \textsuperscript{\rm 2}Ben-Gurion University of the Negev\\
    babramow@tulane.edu, 
    omerlev@bgu.ac.il, 
    nsmattei@tulane.edu
}

\newtheorem{proposition}{Proposition}

\newtheorem{observation}{Observation}

\newtheorem{corollary}{Corollary}

\newtheorem{theorem}{Theorem}

\newtheorem{definition}{Definition}

\newtheorem{example}{Example}

\newcommand{\judge}{\ensuremath{j} }
\newcommand{\judges}{\ensuremath{J} }
\newcommand{\experts}{\ensuremath{E} }
\newcommand{\expert}{\ensuremath{e} }
\newcommand{\vote}{\ensuremath{v} }

\usepackage[textwidth=1.2cm]{todonotes}
\definecolor{kentuckyblue}{RGB}{0, 93, 170}

\begin{document}

\maketitle

\begin{abstract}
We consider the problem of determining a binary ground truth using advice from a group of independent reviewers (experts) who express their guess about a ground truth correctly with some independent probability (competence) $p_i$. In this setting, when all reviewers are competent with $p \geq 0.5$, the Condorcet Jury Theorem tells us that adding more reviewers increases the overall accuracy, and if all $p_i$'s are known, then there exists an optimal weighting of the reviewers.

However, in practical settings, reviewers may be noisy or incompetent, i.e., $p_i \leq 0.5$, and the number of experts may be small, so the asymptotic Condorcet Jury Theorem is not practically relevant. In such cases we explore appointing one or more chairs (judges) who determine the weight of each reviewer for aggregation, creating multiple levels. However, these chairs may be unable to correctly identify the competence of the reviewers they oversee, and therefore unable to compute the optimal weighting.

We give conditions when a set of chairs is able to weight the reviewers optimally, and depending on the competence distribution of the agents, give results about when it is better to have more chairs or more reviewers. Through numerical simulations we show that in some cases it is better to have more chairs, but in many cases it is better to have more reviewers.
\end{abstract}

\section{Introduction}
People have been struggling with finding the \emph{correct} answer for millennia\footnote{Fans of \emph{The Hitchhiker's Guide to the Galaxy} know it is \textbf{42}.}. %\nickin{Omer -- I think we get one joke not 2 :-)} 
In ancient times, when faced with a problem that required discovering a ground truth, two main approaches dominated. The first, less common today, was to approach deities and either ask them to intervene on the randomness of the world (as in the Book of Joshua, Chapter 7), which is a bit akin to sortition \cite{FGGHP21}; or to ask the deity's wisdom directly (e.g., the Oracle at Delphi). The second approach, still in widespread use today, is to try to assess the known information and draw a conclusion. This can either be done by laymen, the basic premise of the jury system as established by Magna Carta, or by people with expertise. In both cases, groups of people are used (instead of single individuals) to increase the reliability and accuracy of the answers, building on the ``wisdom of the crowds''.
%'the mathematical foundation of the Condorcet Jury Theorem \cite{de1785essai,sep-jury-theorems} leading to the ``wisdom of the crowds''.

Mathematical analysis of using a group of agents -- a \emph{jury}, or a set of \emph{experts} -- to assess information and make a decision has been done since at least Condorcet's time, in the late 18th century, when he established the Condorcet Jury Theorem \cite{de1785essai,sep-jury-theorems}. In the standard jury setting, agents vote on a binary ground truth and the objective is to aggregate their votes, using a voting rule, to maximize the probability of the outcome being correct. In this setting it is typical to assume that agents guess the ground truth correctly with some independent probability (competence) $p_i$, we call agents competent when $p > 0.5$ and incompetent when $p \leq 0.5$\footnote{The term competent is not meant to express a value judgment.}. 
% \footnote{We use the term incompetent for clarity and do not pass any kind of moral judgement on these agents, as they may be wrong for good reasons!} 
% For example, a remote sensor may have drifted so far off its initial calibration to be \emph{reliably} wrong, as has happened with many spacecraft over the years \cite{bar2003effect}.}
According to the Condorcet Jury Theorem (CJT), when the agents are competent, the collective accuracy of their majority vote tends to correctness as the number of agents increases. Even with a relatively small number of highly competent agents, accuracy can be very high. However, this result, which basically tells us that groups are less prone to mistake than individuals, rests on a knife's edge. If the agents are even minimally incompetent then, as the population grows, their collective accuracy under majority voting tends to 0, and a small group of highly incompetent agents stands no chance.

In the world around us, this idea is used everywhere -- in judicial settings (juries), in academic conferences (peer evaluation), in voting for political leaders or in referendums, and even in settings with inanimate agents, such as aggregating sensor outputs into a single reading or indicator.

The precariousness of the Condorcet Jury Theorem stems from the underlying aggregation procedure, majority voting, being anonymous, thus treating all agents equally, regardless of their competence. When agent competences can be different, majority rule is generally sub-optimal, and if one knows exactly the agents' level of expertise, the optimal aggregation method for maximizing accuracy with any number of independent experts and any competences is to use a weighted majority rule in which each agent's weight is the log-odds of their competence~\cite{shapley1984optimizing,nitzan1982optimal}. Somewhat surprisingly, the optimal weight of each agent does not depend on the competences of the other agents or even on the total number of agents. However, the assumption that the competence of each agent is exactly known by others is highly unrealistic.

We consider a variant of the classic jury setting, inspired by the domain of academic peer review \cite{Shah22CACM}, which attempts to address these issues. Since the quality of reviewers may not be known by the conference Program Chairs, many conferences (e.g., AAAI) appoint more senior researchers as SPCs (or chairs) to evaluate the reviewers and decide on how to aggregate their views. Such a multi-level process inspired our model: There are not only reviewers, who we will call \emph{experts}, but also chairs, who we will call \emph{judges}, that evaluate the experts and assign them weights.

Analyzing this setting is particularly interesting when the number of agents is relatively small, and therefore we cannot rely on the asymptotic guarantees of the CJT; as well as when there is a potential for significant deviation among agent competency, even when the particular competence values are unknown; or when agents can be incompetent, i.e., they will make the wrong decision most of the time\footnote{In academic peer evaluation it is uncommon for reviewers to be given negative weights, as is required for the log-odds rule. But it is common in other settings, e.g., sensors or proxy voting scenarios where one might want to always do the opposite of a political rival.}. We examine when such a two-level system works well, under what conditions it might be worthwhile to implement it, and when is it better to have an expert become a judge.

\paragraph{Contribution}
We propose and investigate a model of multi-level jury problems for use when we have a small number of possibly unreliable agents. We show that when we know the agent competences exactly (or even approximately), we can find an optimal aggregation procedure, as long as the judges are competent. When the agents' (experts and judges) competences are unknown, we provide a set of numerical experiments demonstrating that adding more than a single judge is rarely helpful, and, indeed, in some cases, the potential damage of a less competent judge is enough to prefer to avoid judges completely.

% \smallskip
% \noindent
% \textbf{Contribution.}
% We propose and investigate a model of multi-level jury problems for use when we have a small number of possibly unreliable agents. We show that when we know the agent competences exactly, we can find an optimal aggregation procedure. When we do not know the competences exactly, we provide a set of numerical experiments demonstrating that in many cases one is better of only appointing a single chair to weight the agents.\nick{come back and make sure this is punchy..}

\section{Related Work}
%Our focus on imperfect judges is inspired by work on academic peer review \cite{Shah22CACM}. In peer review, conference Program Chairs typically invite reviewers whose reviewing quality may not be known. Many conferences (e.g., AAAI) appoint more senior researchers as SPCs (or chairs), to aggregate the reviews, taking into account the reliability of the reviewer. This multi-level process inspired our model.
%
There is a long history of studying the Condorcet jury model and its extensions (e.g., \citet{berend1998condorcet,ben2000nonasymptotic,grofman1978judgmental,feld1984accuracy}), which appears in many areas, including computer science, philosophy, and economics. Our investigation is primarily based off the literature on weighting experts in both the offline~\cite{shapley1984optimizing,nitzan1982optimal} and online settings~\cite{cesa1997use,vovk1990aggregating,freeman2020no, berend2015finite}, in this work we restrict our focus to a single decision. The overall CJT model can be seen in portfolio solver techniques where slower, more reliable algorithms evaluate ensembles of faster, less accurate algorithms \cite{thornton2013auto} and in boosting techniques from machine learning, where one aggregates weakly competent classifiers into a better overall classifier \cite{schapire2013boosting}.

In settings with repeated decisions, the competences of the experts can be estimated based on their voting history. Their competence might be estimated by their similarity to other agents, of how often they agreed with the decision outcomes in the past~\cite{grofman1983thirteen,baharad2012beyond,romeijn2011learning}. In our setting, however, we do not have access to this history and cannot use it to estimate competences. Indeed, in peer review, one may have a notion of other reviewers' competency, but rarely does one co-review with another to form a precise estimate.% from this data alone.

Our emphasis on imperfect judges is also inspired by work on ``wisdom of the crowds" and crowdsourcing~\cite{surowiecki2005wisdom,brabham2013using,brabham2015crowdsourcing}, proxy voting~\cite{abramowitz2019flexible,pivato2020weighted} and truth-tracking in Liquid Democracy~\cite{zhang2022tracking,becker2021unveiling}. But, as noted above, a major inspiration has been the work on academic peer evaluation \cite{Shah22CACM}, in which experts assess each other's competences. Some treat this matrix as a Markov chain, and use its eigenvector values as the experts' weights~\cite{grofman1983determining} in a manner reminiscent of some peer-evaluation models \cite{page1999pagerank,Wal14,LMTZ23}. 
This contrasts with our setting in which the set of agents who vote and the set of agents who weight the voters are disjoint.
 
Finally, the problem of partitioning agents into judges and experts is also related to the problem of computing optimal committee sizes~\cite{magdon2018mathematical,revel2021optimal}. 
There has also been attention paid to how group accuracy depends on the size of the group and their mean competence \cite{grofman1978judgmental,grofman1984group}, which is reflected, in part, in our simulations.

\section{Model and Notation}
Our model has two types of agents; judges and experts. 
Let \experts be a set of $m$ experts and \judges be a set of $n$ judges.
The experts vote on a single binary issue where there is only one correct (ground truth) outcome. Without loss of generality, let the options be represented as $\{1,0\}$ where $1$ is correct and $0$ is incorrect. 
Each expert $\expert \in \experts$ has a \emph{competence}, or probability $p_\expert$ of voting correctly, independent of all other experts.
We associate each expert's index with their vote, so expert $\expert \in \experts$ casts a vote $\vote_\expert \in \{1,0\}$ with competence $p_\expert = P(\vote_\expert = 1)$. 
If an agent's competence is above $\nicefrac{1}{2}$ we will say that they are competent, and call them incompetent otherwise.
We assume no one is always correct or always incorrect, and so $0 < p_\expert < 1$.

\vspace{0.2cm}
\noindent
\textbf{Weighted Majority Rules.\;}
For any competence vector and aggregation rule we refer to the probability of producing the correct outcome as the \emph{accuracy}, and reserve the term \emph{competence} to refer to individual agents' probabilities of voting correctly; i.e. accuracy means collective competence.

\begin{definition}[Weighted Majority Rule]
A weighted majority rule gives each expert $\expert \in \experts$ a weight $w_\expert \in \mathbb{R}$ and selects $1$ as the winner if $\sum\limits_{\vote_\expert = 1} w_\expert > \sum\limits_{\vote_\expert = 0} w_\expert$, selects $0$ as the winner if $\sum\limits_{\vote_\expert = 1} w_\expert < \sum\limits_{\vote_\expert = 0} w_\expert$, and uses a tie-breaking rule (e.g. coin flip) for the edge case where these sums are equal.
\end{definition}

\begin{definition}[Simple Majority Rule]
Simple majority rule refers to the weighted majority rule where all weights are equal and positive and ties are broken randomly.
\end{definition}

The Condorcet Jury Theorem tells us that if $p_e \geq 0.5 + \epsilon$ for some $\epsilon > 0$ for all experts, then with simple majority accuracy tends to 1 asymptotically as the number of experts tends to infinity.
%
% The Condorcet Jury Theorem (CJT) tells us that when $p_e \geq 0.5 + \epsilon$ for $\epsilon > 0$ for all experts, accuracy tends to 1 in the infinite limit as we increase the number of experts \cite{de1785essai}.
%
%
A weighting function maps vectors of values in $(0,1)$ (i.e. competences) to equal length vectors of real values.
For any set of experts, including incompetent ones, the optimal aggregation method of experts' votes, is to apply the log-odds weighting function to the experts' competences and use the corresponding weighted majority rule~\cite{shapley1984optimizing,nitzan1982optimal}.

\begin{definition}[Log-Odds Weighting Function]
Given a vector of values in the open unit interval $\vec{p} = (p_1, \ldots, p_m)$, the log-odds weighting function returns the vector $\vec{w} = (w_1, \ldots, w_m)$ where $w_\expert = \log(\frac{p_\expert}{1 - p_\expert})$ for all $1 \leq \expert \leq m$.
\end{definition}

Any weighting of the agents implies a collection of winning coalitions -- subsets of agents who, if they all vote the same way, determine the outcome regardless of the votes of the remaining agents~\cite{taylor1992characterization}. Different weightings may yield the same rule because they imply the exact same winning coalitions. For example, with 5 agents there are exactly 7 distinct weighted majority rules~\cite{karotkin1988essential,karotkin1994variability}. Multiplying the weights of all agents by a constant does not change the winning coalitions and therefore does not change the rule. Similarly, perturbing agent weights by small amounts may not change the winning coalitions. Therefore, while the weights may vary continuously, the accuracy under various weightings will change in discrete steps. In practice, weights may be finite precision rather than true real numbers, and this is also the case in our simulations that use floating point arithmetic, but as long as the rounding tends to be to small to change winning coalitions for most instances its effect will be negligible.

The log-odds weighting rule assigns a positive weight to competent experts when $p_e > 0.5$, weight of zero if $p_\expert = 0.5$, and negative weight to any incompetent expert with $p_\expert < 0.5$. In some settings it may be inappropriate to allow negative weights and better to assume any such weights are rounded up to zero. Bounding weights below by zero has the effect of ignoring the incompetent experts and is therefore qualitatively similar to assuming all experts are competent, though with a smaller number of experts. We therefore focus on the more informative setting where weights can be negative. Negative weights also have real-world motivation. A remote sensor may have drifted so far off its initial calibration to be \emph{reliably} wrong, as has happened with many spacecraft \cite{bar2003effect}. However, we would like to believe that peer reviewers, jurors, and the like are not so reliably wrong that negative weights would be needed.

\vspace{0.2cm}
\noindent
\textbf{Multi-Level Jury Problems.\;}
Each judge $\judge \in \judges$ estimates the competence of each expert $\expert \in \experts$ as $p_{\judge \expert} \in (0,1)$ and assigns them a weight $w_{\judge\expert}$ based on this estimate.
When assigning weights to the experts, our judges always use the log-odds weighting function on their competence estimates.
Intuitively, our model's judges are trying to implement the optimal rule using their estimates of the experts' competences.
Formally, judge $\judge$ assigns expert $\expert$ a weight $w_{\judge\expert} = \log(\frac{p_{\judge\expert}}{1-p_{\judge\expert}})$.
When there are multiple judges, the weight of an expert will be the average weight assigned to them by the judges $w_\expert = \frac{1}{n} \sum\limits_{\judge} w_{\judge \expert}$.
%
%Note that multiplying all expert weights by a positive constant factor does not change the weighted majority rule, so one could equivalently use the sum of the $w_{\judge \expert}$ values, or multiply the weights by any other positive constant.\footnote{The sum of weights rather than the average is more commonly used in the literature on delegative voting, but in our setting the two are equivalent.}

\smallskip
\noindent
\textbf{Perceived Competences.\;}%
Our theoretical results depend on judges using the log-odds weighting but do not depend on how the judges \emph{form} their competence estimates $p_{\judge \expert}$.
To perform empirical analysis we must make assumptions about where these estimates come from.
Rather than drawing the estimates $p_{\judge \expert}$ from some named distribution with mean $p_\expert$, we take an approach inspired by peer review, and assume that the judges are fundamentally similar to the experts in the same way chairs are similar to reviewers.
Each judge $j$ has competence $p_j$ just like the experts and estimates competence of expert $\expert$ as $p_{\judge \expert} = (p_\judge \cdot p_\expert) + (1 - p_\judge)(1 - p_\expert)$, i.e., the probability that expert $\expert$ agrees with them. 
%
%This value is chosen because it corresponds to how often the judge and expert expect to agree. 
When $p_{\judge \expert}$ is derived in this way, we will refer to it as the judge's \emph{perceived competence} of the expert.
%
%While we do not consider voting over a sequence of issues, we can think of this estimation 
As with peer review, a judge may estimate an expert's competency from knowing them professionally, but may not have observed many or any of their past reviews. A judge could also reach this estimate of competency if they observe enough votes from the expert but the ground truth is never revealed, as is the case in some peer prediction settings \cite{witkowski2012peer}.

\begin{example}\label{example:motivation}
Suppose we have 5 experts with competences $\vec{p}_\experts = (0.6, 0.6, 0.6, 0.7, 0.9)$. The optimal log-odds weighting is approximately $\vec{w}^*_\experts = (0.41, 0.41, 0.41, 0.85, 2.2)$. With these weights the most competent expert $(p_\expert = 0.9)$ receives a weight $(w_\expert = 2.2)$ that makes them a dictator in a weighted majority vote, since their weight is greater than all other experts combined. Hence, the accuracy under the log-odds weighting is exactly 0.9. If instead we use simple majority, the accuracy drops to 0.82.
%If all the experts are weighted equally, the accuracy of the weighted majority vote decreases to 0.82.

A judge with competence 0.6 would assign the experts weights of approximately $\vec{w}^{0.6}_\experts = (0.08, 0.08, 0.08, 0.16, 0.323)$ using the log-odds weighting of perceived competences. Note that the fifth expert is no longer a dictator.
%
%is is not equivalent to the optimal weighted majority rule because the first four experts together outweigh the fifth expert alone.
%
How high of a competence would the judge need to have to assign weights that results in the optimal weighting?
%that yield the same dictatorial weighted majority rule as the optimal weighting? 
The judge's competence would have to be greater than $0.962$; a far cry from $0.6$ and higher than all the experts.
And yet, the judge's sub-optimal weighting yields an accuracy of 0.898, which is a great improvement over simple majority, and extremely close to optimal at 0.9!
\Cref{example:motivation} is illustrated in \Cref{fig:example} where we plot the accuracy, sweeping $p_j$ from 0.0 to 1.0.

\begin{figure}[h]
 \centering
 \includegraphics[height=3.5cm]{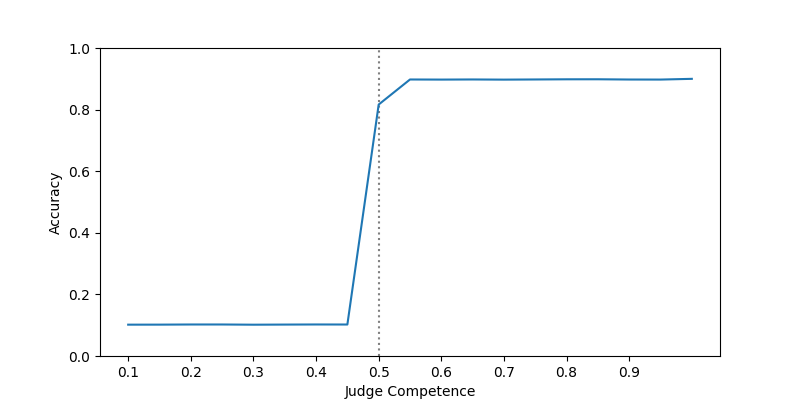}
 \caption{Accuracy of perceived optimal weightings from a single judge with the expert competences as in \Cref{example:motivation}.}
 \label{fig:example}
\end{figure}
\end{example}

In \Cref{example:motivation} we rounded the values of the weights to two decimal places, which did not change the rule implemented. Similarly, whether judges are human, sensors, or algorithms, they do not (always) need to provide high precision of weights. More specifically, the smaller the number of agents, the less chance there is for small perturbations or rounding of the weights to change the rule. This is also why \Cref{fig:example} will be piecewise linear regardless of the step size we choose for the judge's competence.

\section{Optimality and Robustness}\label{sec:opt_robust}
With a single judge, if $p_{\judge \expert} = p_\expert$ for all $\expert \in \experts$, then all experts receive their optimal weight. As noted, small perturbations to the weights do not change the rule because the winning coalitions determined by the weights do not change. Thus, if $p_{\judge \expert}$ is close enough to $p_\expert$ for all experts, they will still produce the optimal weighting. We now establish sufficient conditions for an ensemble of judges to produce the optimal weighting, and provide a condition under which the difference between an expert's weight and their optimal weight tends to be small.
%
% Another corner case of optimality occurs when all experts have equal competence, and the judge happens to assign them all the same weight, even if their estimate of all the experts' competences is incorrect, as long as the inaccuracy of the estimate does not change the sign of the weights.

\Cref{theorem:optimal} shows that when the geometric mean of the judges' perceived competence of experts odds is their true competence odds, 
%i.e., $\left( \frac{p_\expert}{1-p_\expert} \right) = \left(\prod\limits_\judge \frac{p_{\judge\expert}}{1-p_{\judge\expert}} \right)^\frac{1}{n}$, 
all experts are assigned their optimal weights, $w_\expert = w^*_\expert$. This does not require individual judges to know the experts' true competences, and does not depend on the number of experts nor the number of judges.

\begin{proposition}\label{theorem:optimal}
If each judge uses the log-odds weighting function on their estimates of expert competences, and the geometric mean of the judges' estimates of each expert's competence odds is the expert's true odds, then the weighted majority rule using the judges' average weights to weight each expert is exactly the optimal weighted majority rule.
\end{proposition}

\begin{proof}
Since judge $\judge$ gives each expert a weight of $w_{\judge\expert} = \log ( \frac{p_{\judge\expert}}{1-p_{\judge\expert}} )$,
\scriptsize
\begin{align*}
 w_\expert = &\frac{1}{n} \sum\limits_\judge w_{\judge\expert} = \frac{1}{n} \sum\limits_\judge \log ( \frac{p_{\judge\expert}}{1-p_{\judge\expert}} )=\\
 = &\frac{1}{n} \log ( \prod_\judge \frac{p_{\judge\expert}}{1-p_{\judge\expert}} )= \log ( ( \prod_\judge \frac{p_{\judge\expert}}{1-p_{\judge\expert}} )^\frac{1}{n} ) 
\end{align*}
\normalsize
We assume the geometric mean of judge' estimates of the experts' competence odds is correct, i.e., $( \frac{p_\expert}{1-p_\expert}) = (\prod\limits_\judge \frac{p_{\judge\expert}}{1-p_{\judge\expert}} )^\frac{1}{n}$. Thus, $w_\expert = \log( \frac{p_\expert}{1-p_\expert}) = w^*_\expert$.
\end{proof}

% If we take any weighted majority rule and scale all the weights by a positive multiplicative factor, the rule remains the same. Hence, \Cref{theorem:optimal} can be immediately generalized as follows.

% \begin{corollary}\label{corollary:generaloptimal}
% If there is a $\gamma>0$ such that for all $\expert \in \experts$, $\left( \frac{p_\expert}{1-p_\expert} \right) = \left(\prod\limits_\judge \frac{p_{\judge\expert}}{1-p_{\judge\expert}} \right)^\gamma$, then the weights constitute the optimal weighted majority rule.
% \end{corollary}

This result requires judges' competence estimates that must hold for \emph{all} experts. However, suppose there are errors in these collective competence estimates. We want to know how sensitive the weight of a single expert is to such errors. 

\begin{corollary}\label{corollary:optimal}
If the geometric mean of judge estimates of competence is off by some multiplicative factor $\alpha$ for some expert, then the error of that expert's weight is only $\log(\alpha)$.
\end{corollary}

\begin{proof}
In the proof above, assume instead that $\alpha \left( \frac{p_\expert}{1-p_\expert} \right) = (\prod\limits_\judge \frac{p_{\judge\expert}}{1-p_{\judge\expert}} )^\frac{1}{n}$. Then
%
% \begin{align}
 $w_\expert = \log ( \alpha \cdot \frac{p_\expert}{1-p_\expert}) = w^*_\expert + \log(\alpha)$.
% \end{align}
\end{proof}

%(asymptotically there is a Central Limit Theorem indicating that the geometric mean will get closer and closer to the correct value)\omer{I think the parenthesis is true. Need to verify}.\ben{I think we want to avoid asymptotic arguments because we focus on small numbers of voters overall.}

Admittedly, it is not clear in what settings the conditions of \Cref{theorem:optimal} should be expected to hold. Neither can we make claims about what multiplicative factors are realistic in \Cref{corollary:optimal}. But ultimately, what we care about most is the sensitive of the accuracy to errors in competence estimates, which has to do with the set of winning coalitions induced by the weights, not the sensitivity of the weights themselves, although the sensitivity of the weights gives some intuition.

Looking back at \Cref{example:motivation}, we see that %with a single judge using perceived competences, 
for all $p_j > 0.55$ the accuracy rivals that of the optimal rule, with a nearly imperceptible difference. The effect that dominates \Cref{fig:example} is when we move from $p_j < 0.5$ to $p_j = 0.5$ (when the rule becomes simple majority), with another slight bump with a move to $p_j > 0.55$.% There is an asymmetry here, because at $p_j = 0.5$ the rule becomes simple majority, and all experts are competent, so we should not get an accuracy of $0.5$ when $p_j = 0.5$. 

Recent work \cite{baharad2022one} shows that when expert competences are drawn from certain distributions over the range $(\nicefrac{1}{2}, 1)$, simple majority achieves an accuracy close to optimal.
However, as one might expect, when experts can be incompetent the majority rule is no longer a good approximation to the optimal weighted majority rule.
Thus, if judges can at least differentiate the competent from incompetent experts, the weighting they produce should be expected to outperform simple majority rule when there are incompetent agents. In our model, any minimally competent judge with $p_j > 0.5$ is able to achieve this.
We will discuss this more in the next section with an array of experiments with a single judge, but for now we introduce some basic theoretical observations that help us understand the phenomenon.

 \begin{example}[Two Experts]\label{example:two_experts}
 Suppose there are two experts with competences $(p_1, p_2)$ such that $p_1 > p_2$. If $p_2 > \nicefrac{1}{2}$, i.e., both experts are competent, the optimal aggregation rule is to make $p_1$ dictator. However, if $p_1 > \nicefrac{1}{2} > p_2$ then the optimal rule is either to make $p_1$ a dictator if $p_1 \geq 1 - p_2$, or else make $p_2$ an anti-dictator using negative weight such that the outcome is the opposite of however $p_2$ votes. If $\nicefrac{1}{2} > p_1 > p_2$, then the optimal rule makes $p_2$ the anti-dictator symmetrically with the first case.
 \end{example}

 From \Cref{example:two_experts}, we see that even with only two experts, if a judge can determine which experts are competent, and order their competences correctly, this is enough information to produce the optimal rule. With more experts, the situation is more complicated, but our experiments reveal that merely separating the competent experts from incompetent ones creates a large improvement in overall accuracy. Any chair with $p_j > \nicefrac{1}{2}$ using log-odds weightings of the perceived competences can achieve this improvement.

\begin{proposition}[Correct Sign]\label{prop:single_j_correct_sign}
 If sign$(p_{\judge \expert} - 0.5) = $ sign$(p_{\expert}-0.5)$, then sign$(w_{\judge \expert}) = $ sign$(w^*_{\judge \expert})$.
\end{proposition}

\begin{proof}
 \Cref{prop:single_j_correct_sign} follows directly from the fact that $\frac{p}{1-p} > 1$ if and only if $p > \nicefrac{1}{2}$, and therefore $\log(\frac{p}{1-p}) > 0$ if and only if $p > \nicefrac{1}{2}$. Symmetrically for $p < \nicefrac{1}{2}$.
\end{proof}

\begin{proposition}[Correct Order]\label{prop:single_j_correct_order}
 If $p_{\judge \expert}$ is a strictly monotonic increasing function of $p_\expert$, then the order of expert weights given by judge $j$ is the order of the experts' competences. 
\end{proposition}

\Cref{prop:single_j_correct_order} follows from the monotonicity of the log-odds weighting. When a single judge applies the log-odds weighting to their perceived competences of a small set of experts then we can make the following observations.
 
\begin{observation}\label{observation:extremes}
 If $p_\judge = \nicefrac{1}{2}$ then all experts are equally weighed. %$p_{\judge \expert} = \nicefrac{1}{2}$ for all values of $p_\expert$ and 
 If $p_\judge = 1$ then experts are optimal weighed.
 %$p_{\judge \expert} = p_\expert$ for all values of $p_\expert$.
 \end{observation}
 
If $p_\judge = \nicefrac{1}{2}$ then the judge will perceive all experts as having a competence of $\nicefrac{1}{2}$, and therefore assign them the weight of 0, which we treat as giving them equal weight. %a weight of zero, equivalent to . Whenever all reviewers weights are zero, this is treated as an equal weighting of the experts, as if all their weights were 1 and hence the simple majority rule applies.
When $p_\judge = 1$, the judge knows exact competences of the experts and therefore assign them their optimal weights.
 
\begin{observation}\label{observation:monotonic}
 If $p_\judge > \nicefrac{1}{2}$, then $p_{\judge \expert} > p_{\judge \expert'}$ iff $p_\expert > p_{\expert'}$.
\end{observation}
 
This means that a judge's perceived competences of the experts preserves the order of their true competences if $p_j > \nicefrac{1}{2}$. 
When weights are based on perceived competences, this means whenever $p_\judge > 0.5$, the judge will assign all experts' weights with the correct sign and in the correct order. This is because when $p_\judge > 0.5$, $p_{\judge \expert}$ is monotonically increasing in $p_\expert$ and $p_{\judge \expert} > 0.5$ iff $p_\expert > 0.5$. Thus, even a single barely competent judge might give us an edge over simple majority.

\begin{theorem}[Minimal Competent Single Judge]\label{thm:minimal_comp}
 If $p_j > 0.5$ and the judge assigns experts weights according to their perceived competences, the weights given by the judge will have the correct sign and the correct order.
\end{theorem}

 \begin{proof}
 Let $p_\judge = \nicefrac{1}{2} + \varepsilon_\judge$ and $p_\expert = \nicefrac{1}{2} + \varepsilon_\expert$.
\small
 \begin{align*}
 p_{\judge \expert} & = \left(\nicefrac{1}{2} + \varepsilon_\judge \right) \left(\nicefrac{1}{2} + \varepsilon_\expert \right) + \left(\nicefrac{1}{2} - \varepsilon_\judge \right) \left(\nicefrac{1}{2} - \varepsilon_\expert \right)\\
 % & = \frac{1}{4} + \nicefrac{1}{2}\left( \varepsilon_\judge + \varepsilon_\expert\right) + \varepsilon_\judge \varepsilon_\expert + \frac{1}{4} - \nicefrac{1}{2}\left( \varepsilon_\judge + \varepsilon_\expert\right) + \varepsilon_\judge \varepsilon_\expert\\
 & = \nicefrac{1}{2} + 2 \varepsilon_\judge \varepsilon_\expert \label{Eq:11}
 \end{align*}
\normalsize
 If $\varepsilon_\judge > 0$ and $\varepsilon_\expert > 0$, then this value is greater than $\nicefrac{1}{2}$, if $\varepsilon_\judge > 0$ and $\varepsilon_\expert < 0$, then this value is less than $\nicefrac{1}{2}$, and if $\varepsilon_\expert = 0$ then this value is exactly $\nicefrac{1}{2}$. The proposition then follows from \Cref{prop:single_j_correct_order} and \Cref{prop:single_j_correct_sign}.
 \end{proof}

\begin{figure*}[t]
\centering
\begin{subfigure}{0.3\textwidth}
 \includegraphics[width=\textwidth]{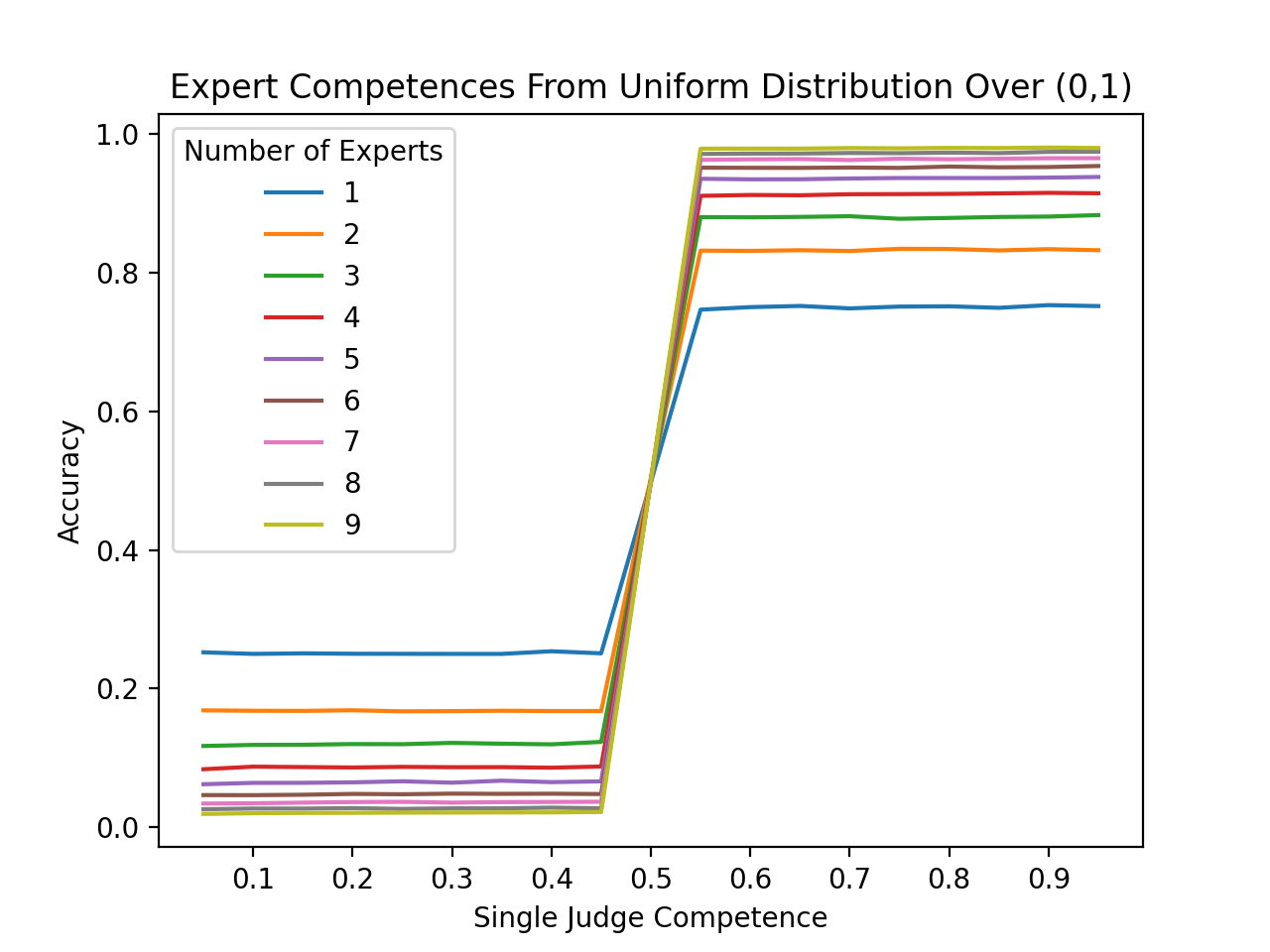}
 \caption{Uniform distribution over $(0,1)$}
\label{fig:E1_uniform_01}
\end{subfigure}
\hfill
\begin{subfigure}{0.3\textwidth}
 \includegraphics[width=\textwidth]{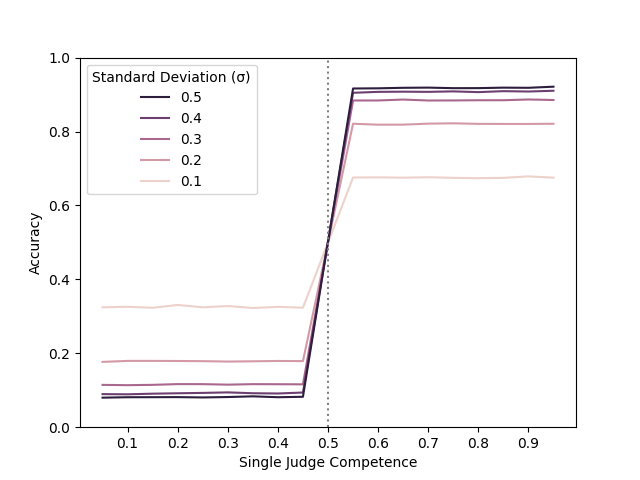}
 \caption{5 experts with truncated (in $(0,1)$) Gaussian competences $\mathcal{N}(\nicefrac{1}{2},\sigma)$}
 \label{fig:E1_gaussian_01_stddevs}
\end{subfigure}
\hfill
\begin{subfigure}{0.3\textwidth}
 \includegraphics[width=\textwidth]{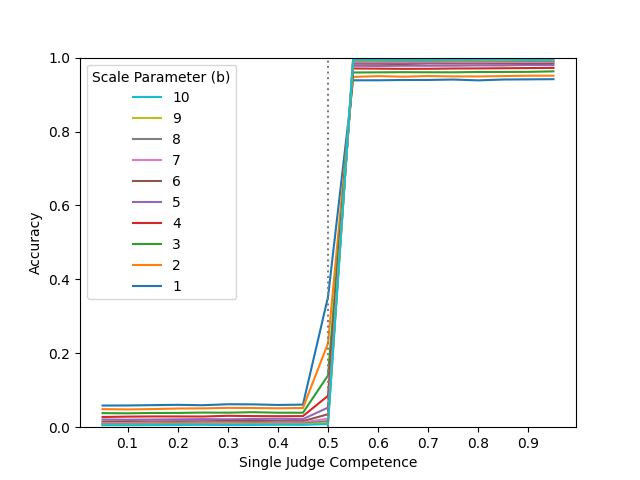}
 \caption{5 experts with exponentially distributed competences $(0,1)$}
\label{fig:E1_exponential_01}
\end{subfigure}
\hfill
 \begin{subfigure}{0.3\textwidth}
 \includegraphics[width=\textwidth]{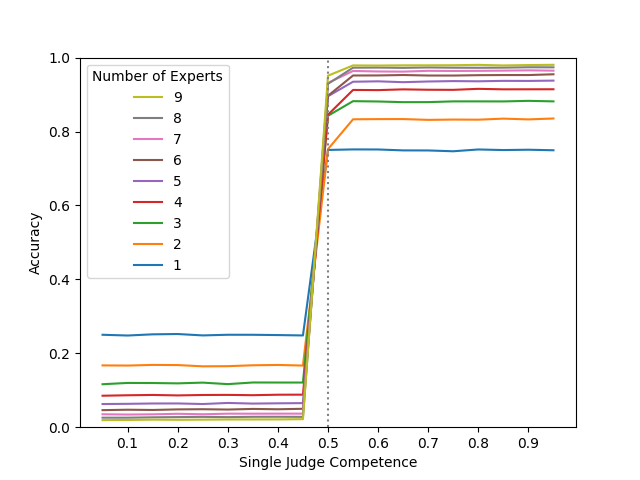}
 \caption{Uniform distribution over $(0.5,1)$}
\label{fig:E1_uniform_05,1}
\end{subfigure}
\hfill
\begin{subfigure}{0.3\textwidth}
 \includegraphics[width=\textwidth]{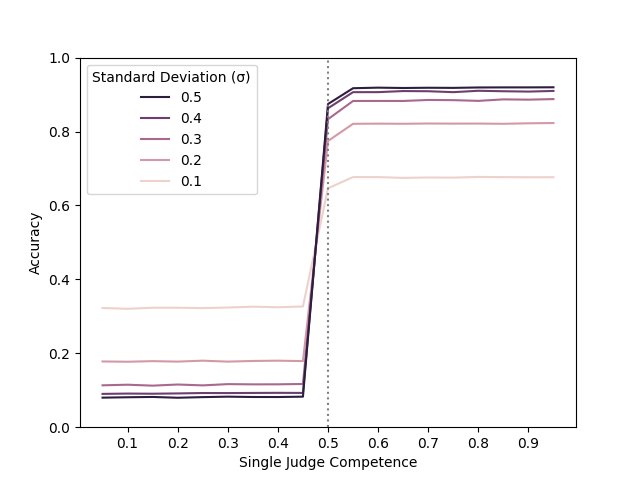}
 \caption{5 experts with truncated (in $(0.5,1)$) Gaussian competences $\mathcal{N}(\nicefrac{1}{2},\sigma)$}
\label{fig:E1_gaussian_05,1_stddevs}
\end{subfigure}
\hfill
\begin{subfigure}{0.3\textwidth}
 \includegraphics[width=\textwidth]{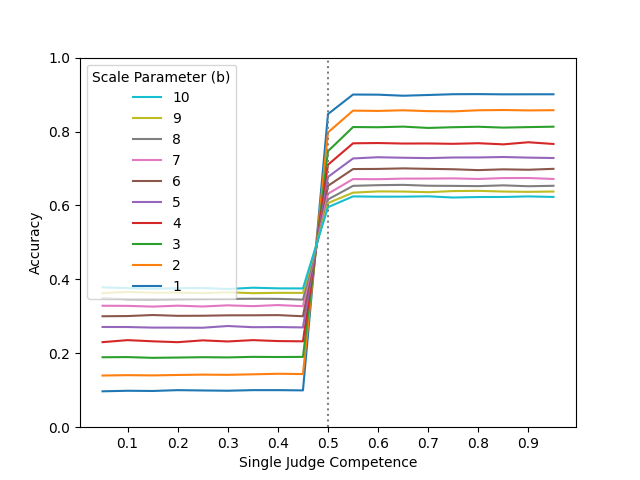}
 \caption{5 experts with exponentially distributed competences in $(0.5,1)$}
 \label{fig:E1_exponential_05,1}
\end{subfigure}
\caption{Accuracy with a single judge and expert competences drawn i.i.d. from a distribution with support $[0.001, 0.999]$ (top row) or support $[0.501, 0.999]$ (bottom row).}
\label{fig:single_judge_(0,1)}
\end{figure*}

We can generalize \Cref{prop:single_j_correct_sign} by replacing the requirement that $p_j > \nicefrac{1}{2}$ with the requirement that the geometric mean of judges' estimated competence odds is greater than 1. Notice that the requirements for this theorem are far weaker than for the optimality demanded by \Cref{theorem:optimal}.

\begin{proposition}[Correct Sign]\label{theorem:rightsign}
If the geometric mean of every expert's estimated competence from the judges is greater than $1$ whenever $p_i > \nicefrac{1}{2}$, less than $1$ whenever $p_i < \nicefrac{1}{2}$, equal to $1$ when $p_i = \nicefrac{1}{2}$, every expert will be assigned a weight with the same sign as their optimal weight.
\end{proposition}

\begin{proof}
%We simply take the inequality that we want to hold and express it in terms of the geometric mean of judges' estimates.
%
Suppose $p_\expert > \nicefrac{1}{2}$. Their optimal weight is positive, and so we need the following to hold: 
\scriptsize
\begin{align*}
 \sum\limits_{\judge \in \judges} \log\left(\frac{p_{\judge\expert}}{1-p_{\judge\expert}}\right) & > 0\\ 
\prod\limits_{\judge \in \judges} \frac{p_{\judge\expert}}{1-p_{\judge\expert}} & > 1 \\ 
\left(\prod\limits_\judge \frac{p_{\judge\expert}}{1-p_{\judge\expert}} \right)^\gamma & > 1
\end{align*}
\normalsize
for $\gamma > 0$. When $\gamma = \frac{1}{n}$ this is the geometric mean.
The case for $p_\expert < \nicefrac{1}{2}$ is symmetric with flipped inequality, and for $p_\expert = \nicefrac{1}{2}$ is the same but with strict equality.
\end{proof}

It is straightforward to see that \Cref{prop:single_j_correct_order} and \Cref{thm:minimal_comp} can similarly be generalized to multiple judges, but we omit these here due to space constraints.

%\ben{We omit the similar generalization to many judges of \Cref{prop:single_j_correct_order} and \Cref{thm:minimal_comp} for space.}

% \ben{Does not imply accuracy is strictly monotonically increasing. But we conjecture that it is weakly monotonically increasing with judge competence.}

\section{Single Judge}
To better understand the behavior we see in \Cref{example:motivation} and \Cref{{sec:opt_robust}}, we undertake a set of numerical experiments to investigate how the accuracy varies with the competence of a single judge.
In our simulations, the experts' competences are drawn i.i.d from various distributions. All experiments were run for 100,000 iterations for each parameterization of the problem instance so that the variances are negligible.

We consider three distributions of expert competences: uniform, truncated normal, and truncated exponential. The uniform distribution reflects settings where the experts can equally have any competencce; the exponential distribution models settings where the expertise tends to be rare% or the product of feedback effects from learning
; and the normal distribution is appropriate for a common expertise, coalescing around a mean value \cite{Tal07}. %\ben{Cite superforecasters..?}\omer{cite Black Swan?}
 
The top row of Figure~\ref{fig:single_judge_(0,1)} shows accuracy as a function of the single judge's competence when expert competences are distributed over the interval $[0.001, 0.999]$ according to the uniform, truncated normal ($\mathcal{N}(\nicefrac{1}{2},\text{varying }\sigma)$), and truncated exponential distributions (using the density function $\frac{e^{-x}}{1-e^{-b}}$ for varying values of $b$) respectively. Only Figures~\ref{fig:E1_uniform_01} and~\ref{fig:E1_gaussian_01_stddevs} exhibit true symmetry because competences are drawn from a symmetric distribution with mean $0.5$, and Figures~\ref{fig:E1_uniform_05,1} and~\ref{fig:E1_gaussian_05,1_stddevs} show behavior most similar to \Cref{fig:example}.% \omer{Decreasing from 0? From 1?}\ben{@Omer, not sure what this means}.

In the top row of Figure~\ref{fig:single_judge_(0,1)} we can have highly incompetent experts, but even in this setting whenever the judge has competence $p_\judge > \nicefrac{1}{2}$, high overall accuracy is achieved. This is because the ability of the judges to differentiate competent experts from incompetent ones is of primary importance, and \Cref{theorem:rightsign} shows that a judge using perceived competences is able to do this.

In Figures~\ref{fig:E1_uniform_01}-\ref{fig:E1_exponential_01}, once the judge passes a minimum threshold of competence, little is gained from increasing $p_j$. % \nickin{say that after 0.6? -- make sure we mark the 0.5 now to make it more clear}
Interestingly, in \Cref{fig:E1_gaussian_01_stddevs} we see that when expert competences are distributed normally with mean $\nicefrac{1}{2}$, higher variance leads to higher collective accuracy. 
This appears to be because a judge with sufficiently high competence can differentiate between highly competent and minimally competent experts, and then leverage the benefits of having highly competent experts when they are present.
% We conjecture that this increase in accuracy is monotonic in the judge's competence for all expert competence vectors. That is, for all vectors of competences $\vec{p}_E$, if $p_j > p_{j'}$, accuracy is higher under the log-odds weighting by $j$ than the log-odds weighting by $j'$. %\nick{break out into conjecture, why don't we prove?} \ben{"For any vector of expert competences..." Make conjecture formal in an environment. Cite the problem of ordering weighted majority rules that has long been open. Nitzan/Paroush}\omer{I agree with Ben. This shouldn't be broken.}
%\nickin{Omer, decide if we even need to say this..if so do we want to say this is an open question and cite? -- take citation from the following parargraph where we had the formal conjecture -- we either should say it with a cite or just don't discuss at all}

In the bottom row of Figure~\ref{fig:single_judge_(0,1)} we show accuracy as a function of the single judge's competence when expert competences are distributed over the interval $[0.501, 0.999]$ according to the uniform, truncated normal, and truncated exponential distributions respectively.
This is closer to prior work in the literature where all experts are assumed to be competent.
Unlike in Figure~\ref{fig:E1_uniform_01} and~\ref{fig:E1_gaussian_01_stddevs}, which were based on symmetrical distributions, we now see a distinctive asymmetry around $p_j = 0.5$. When the judge's competence is $\nicefrac{1}{2}$, the judge gives all experts the same weight, so when experts competence is symmetrical around $\nicefrac{1}{2}$ (as in the upper row of Figure~\ref{fig:single_judge_(0,1)}), 
%\nick{what does any mean here?} 
this results in an accuracy of $\nicefrac{1}{2}$, but when they cannot be incompetent, the resulting accuracy is higher -- almost optimal \cite{baharad2022one}.
% However, when the judge's competence grows beyond $\nicefrac{1}{2}$, the growth The accuracy in this case is no longer $\nicefrac{1}{2}$ as in the previous case (since reviewers could be of any competence, and This appears to be because when $p_\judge = \nicefrac{1}{2}$ all experts' weights are equal, and this tends to be close to optimal \cite{baharad2022one}. Even a small increase in the competence of the judge appears to produce weights whose accuracy is nearly indistinguishable from that of the optimal weights. However, for the incompetent judges, weighting competent experts with negative weights and incompetent experts with positive weights causes a severe drop in accuracy. \omer{I don't get this explanation. What is the difference caused by agents being better, that results in a more gradual slope? It seems the quality \emph{drops} because the experts are better?}
%
In contrast to the top row of Figure~\ref{fig:single_judge_(0,1)}, when all experts are competent, there is a large difference in accuracy for truncated exponential distributions with different scale parameters.
%

%(We have included additional experiments with a single judge in the Appendix).

%\omer{I have commented out Ben's conjecture that Nick asked for, as I think it takes us away from the topic of the paper. What do you think, Nick?}
%Given a vector of expert competences, it is a long-standing open question how to order the possible weighted majority rules from most accurate to least accurate when there are more than 5 experts~\cite{karotkin1988essential, karotkin1994variability}. We conjecture that in our experiments the increase in accuracy we observe is strictly monotonic in the judge's competence for all expert competence vectors. If this conjecture holds true, it may help us better understand how to order weighted majority rules by their accuracy.
%
%\begin{conjecture}
% For all expert competence vectors $\vec{p}_\experts$, if $p_j > p_{j'}$, then accuracy is higher under the log-odds weighting by $j$ than the log-odds weighting by $j'$ using perceived competences.
%\end{conjecture}

\section{Should We Add a Judge or an Expert?}
Empirically, with a single judge the accuracy improves as the judge's competence grows, and we know from the Condorcet Jury Theorems that as the number of experts increases, if the experts are competent, accuracy will increase.
It follows immediately that if the conditions of \Cref{theorem:rightsign} hold, then accuracy will increase as the number of experts increases whether they are competent or incompetent, as long as they have competences that are not equal to $\nicefrac{1}{2}$.

We examine the balance between the benefits of increasing the number of judges and increasing the number of experts. That is, with a fixed set of agents of unknown competences, how should they be partitioned between experts and judges? This problem is faced by any scientific conference with a hierarchical structure: how to divide its Program Committee between reviewers and SPCs, ACs, etc.
\begin{figure*}[t]
\centering
\begin{subfigure}{0.3\textwidth}
 \includegraphics[width=\textwidth]{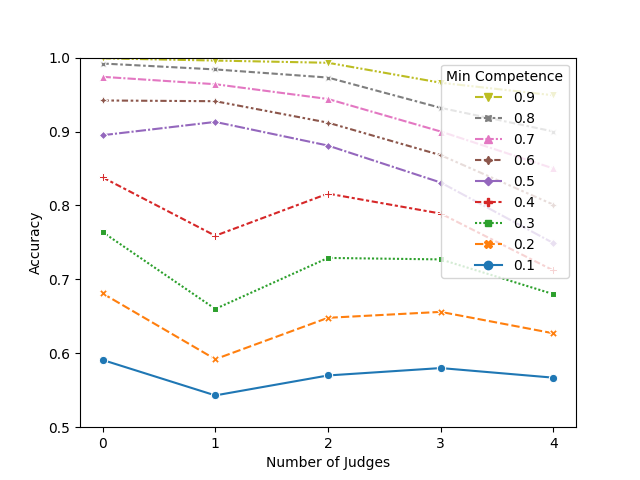}
 \caption{5 agents with competence from uniform distribution over $(\min,1)$}
\label{fig:5_uniform}
\end{subfigure}
\hfill
\begin{subfigure}{0.3\textwidth}
 \includegraphics[width=\textwidth]{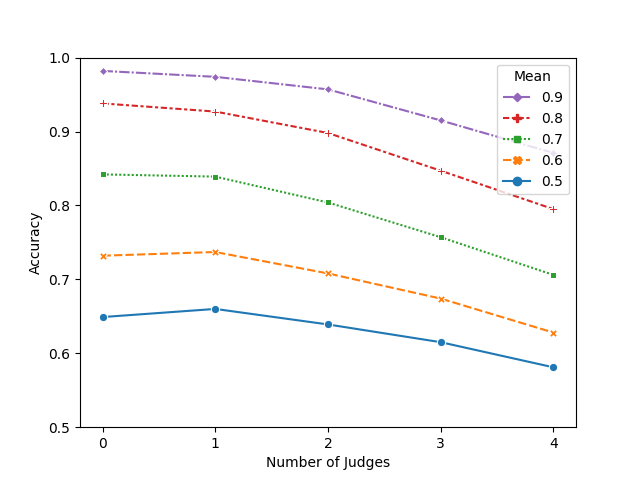}
 \caption{5 agents with truncated Gaussian competences $\mathcal{N}(\text{mean},0.1)$ in $(0,1)$}
 \label{fig:5_gauss}
\end{subfigure}
\hfill
\begin{subfigure}{0.3\textwidth}
 \includegraphics[width=\textwidth]{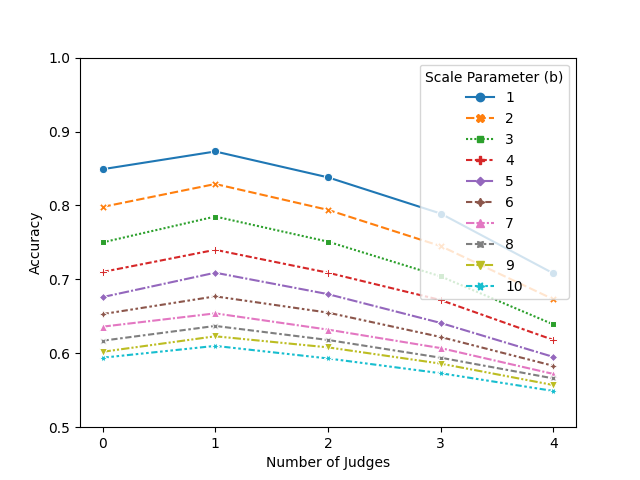}
 \caption{5 agents with exponentially ditributed competences truncated in $(0.5,1)$}
\label{fig:5_exponential}
\end{subfigure}
\hfill
 \begin{subfigure}{0.3\textwidth}
 \includegraphics[width=\textwidth]{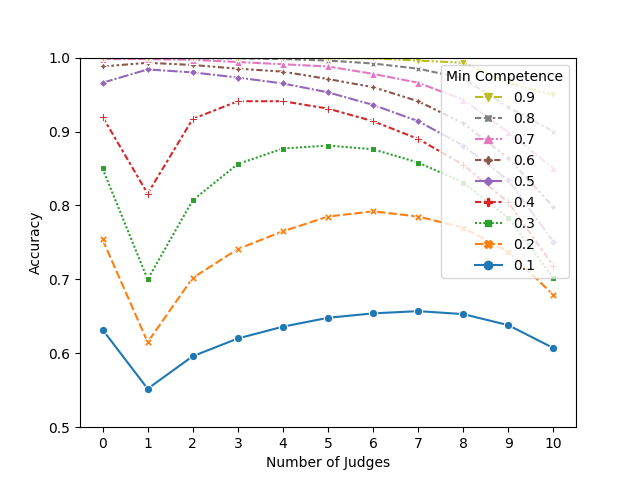}
 \caption{11 agents with competence from uniform distribution over $(\min,1)$}
\label{fig:11_uniform}
\end{subfigure}
\hfill
\begin{subfigure}{0.3\textwidth}
 \includegraphics[width=\textwidth]{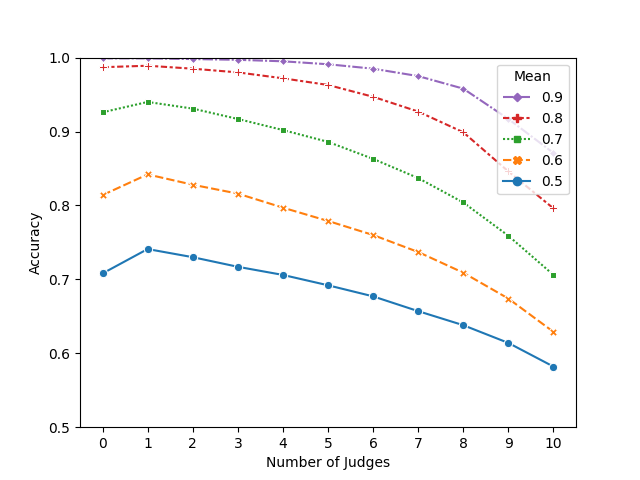}
 \caption{11 agents with truncated Gaussian competences $\mathcal{N}(\text{mean},0.1)$ in $(0,1)$}
 \label{fig:11_gauss}
\end{subfigure}
\hfill
\begin{subfigure}{0.3\textwidth}
 \includegraphics[width=\textwidth]{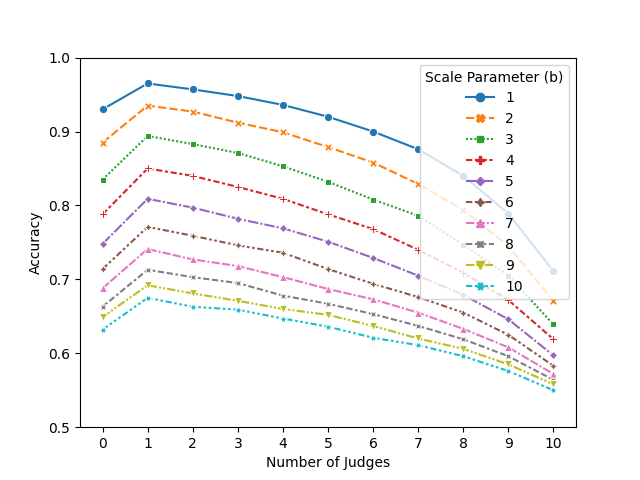}
 \caption{11 agents with exponentially distributed competences truncated in $(0.5,1)$}
\label{fig:11_exponential}
\end{subfigure}
\caption{Accuracy from partitioning a set of agents randomly into judges and experts given that all agent competences come i.i.d. from the same distribution.% Figures left to right use truncated uniform, normal, and exponential distributions. The various plot lines correspond to different distribution parameters; minimum competence for the uniform distributions and scale parameter for the exponential distributions, respectively.
}
\label{fig:partition}
\end{figure*}

We first draw agents' competences from uniform distributions with varying lower bounds and examine the optimal number of agents to set aside as non-voting judges rather than experts when we have 5 and 11 agents, respectively (\Cref{fig:5_uniform,fig:11_uniform}). For both number of agents, we see that \emph{setting aside a single agent as judge diminishes the accuracy compared the simple majority rule in almost all cases}. This is more pronounced when there is a possibility the judge will have competence below $\nicefrac{1}{2}$, i.e., when a lone judge is incompetent they give all competent experts negative weights and incompetent experts positive weights. The only case where a judge is helpful is when the minimum value of agents is $\nicefrac{1}{2}$, perhaps because there is high enough chance that the judge will be helpful, and the agents' competence is not guaranteed to be high enough that losing the judge as an expert is too big a hit. Even adding more judges, at best, returns the accuracy to the level of a simple majority rule, though most commonly it does not. In the 11 agent case, \Cref{fig:11_uniform}, this effect is even more pronounced than in 5 agent case. In the 11 agent case, adding enough judges can eventually bring peak accuracy to slightly above the simple majority, though it requires roughly an even split between judges and experts (or even slightly more judges).

In contrast to the uniform distribution, when drawing competences from the normal  and the exponential distributions, things are a bit different. They show that a single judge can be productive. With normal distributions, when the mean is high but not extremely so (\Cref{fig:5_gauss,fig:11_gauss}), adding a judge helps. When the mean is very high (0.8 and above), aggregating all agents as experts seems to be better than having a judge, for whom there is still a probability of being bad. But when agents are with a lower mean, having a judge seems to help, and this is true even for a mean of 0.5, in which there is a probability of $0.5$ that the judge will be incompetent. This pattern appears for 5 agents, but, as in the uniform case, it is more pronounced for 11 agents.

For the exponential distributions (\Cref{fig:5_exponential,fig:11_exponential}), this property is stronger -- it is \emph{always} beneficial, for our parameters, for agents to have one judge, and that improves over a simple majority. This is likely due to the fact that the small loss of accuracy from having one fewer experts is made up for by the ability of even a minimally competent judge (and all agents are competent in this distribution) to distinguish highly competent experts from less competent ones. Unlike in the uniform case, adding more judges (after a single one) is never helpful compared to a single judge (though sometimes two judges are better than simple majority).

This exploration implies that the division of people in scientific conference is counter-intuitive: the existence of multiple layers above regular experts (e.g., SPC, AC, which are, de facto, multiple judges) does not seem to be helpful. It seems better to have a flat hierarchy (i.e., fewer judges) and use simple majority, despite it being often frowned upon as a strict measure of a paper's quality. We did not investigate the case where a judge is explicitly better than experts, but it is not clear that we are better off in using a judge in such a case, as losing a top expert incurs a cost. Indeed, it is not at all clear that the best judge is the one with the highest competence, and we leave this open question to future research.%\omer{I added this sentence as suggested. Not sure if it highlights too much an issue with the paper…}\nickin{I think it's okay, would keep, it does invite the question but I think it's okay.}%\omer{should we add a sentence acknowledging that we don't examine judges of better competence than experts? NICK-- YES PLEASE!}

\section{Discussion and Future Work}
We consider a multi-level jury problem in which experts are given weights according to estimates from judges of their competence. We focus on settings where there is a small number of agents, so the classic asymptotic results from the literature do not apply, as well as cases where it is possible for agents to be incompetent (i.e., their chance of being correct might be less than $\nicefrac{1}{2}$).

We prove several conditions guaranteeing good outcomes, as well as some which give some minimal guarantees on the quality of the result. Moreover, we show some cases where judges bring a meaningful benefit to the process. However, our results regarding how to divide a group of agents -- a particularly relevant issue for scientific conferences -- indicate that multiple judges may be unhelpful, and there are cases (e.g., uniform distributions) in which an additional expert is more valuable than a judge. %This calls into question decisions made today in various fields, such as conference reviewing.

There are several interesting future directions. One is to reconsider the problem we have presented here when the weights given by experts must all be non-negative, or when it is required for each judge $\judge$ that $\sum_{\expert \in \experts} w_{\judge \expert} = 1$ (as required in \citet{ALMRW16,ALMRW19} for the setting of peer evaluation). Another is to examine what happens when the hierarchy level is increased by adding an additional layers (as in large conferences, which have Area Chairs in charge of SPCs, in charge of PC members). At what point does it no longer become helpful (or begin to be helpful)? Can a guarantee of minimal quality of judges change the value proposition of having them?% \nick{add cite to peer paper for normalized weights}

% \begin{example}
% Consider an autonomous system with two kinds of sensors. Both sensor types take regular measurements of the same kind (e.g. path obstructed or unobstructed). The first type of sensor is cheap, takes multiple measurements each second, and can transmit a single bit every second, but accuracy is highly variable across sensors. The second type of sensor is more costly, takes a measurement every few seconds, and is more reliable, but can only receive and transmit a few bits each minute. If decisions must be made quickly, on the order of seconds, the second sensor might seem useless. However, if these slower sensors can be used to judge the accuracy of the faster sensors at regular intervals, the overall accuracy of the entire ensemble may be improved. The same intuition applies to the use of learning algorithms and approximation algorithms that require different amounts of time to compute in time-sensitive applications. 
% \end{example}

\section{Acknowledgements}
Nicholas Mattei was supported by NSF Awards IIS-RI-2007955, IIS-III-2107505, and IIS-RI-2134857, as well as an IBM Faculty Award and a Google Research Scholar Award. This work was supported by the Tulane University Jurist Center for Artificial Intelligence and the Tulane University Center for Community-Engaged Artificial Intelligence. Ben Abramowitz was supported by the NSF under Grant \#2127309 to the Computing Research Association for the CIFellows Project. Omer Lev was supported, in part, by NSF-BSF grant \#2021659, and by Israel Science Fund (ISF) grants \#1965/20 and \#3152/20.

\bibliography{arxiv}

% \clearpage
% \appendix
% \include{supp}

\end{document}